\documentclass{article}

\usepackage{arxiv}

\usepackage[utf8]{inputenc} 
\usepackage[T1]{fontenc}    
\usepackage{hyperref}       
\usepackage{url}            
\usepackage{booktabs}       
\usepackage{amsfonts}       
\usepackage{nicefrac}       
\usepackage{microtype}      
\usepackage{lipsum}		    
\usepackage{graphicx}
\usepackage[numbers,square]{natbib}
\usepackage{doi}
\usepackage{amsmath}
\usepackage[bottom]{footmisc}
\usepackage{footnote}
\hypersetup{hidelinks}
\usepackage{lipsum}

\title{Harnessing Digital Pathology And Causal Learning\\To Improve Eosinophilic Esophagitis Dietary\\Treatment Assignment}

\makeatletter
\def\@fnsymbol#1{\ensuremath{\ifcase#1\or \dagger\or *\or \ddagger\or
   \mathsection\or \mathparagraph\or \|\or **\or \dagger\dagger
   \or \ddagger\ddagger \else\@ctrerr\fi}}
\makeatother

\date{}

\author{
    \href{https://orcid.org/0000-0003-4591-7018}
    {\includegraphics[scale=0.06]{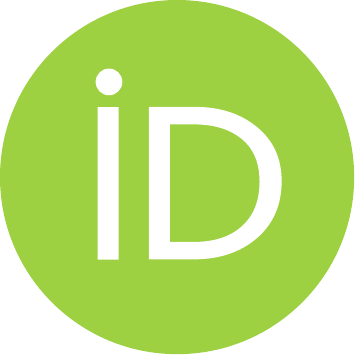}\hspace{1mm}Eliel Aknin$^{1,2,}$ \thanks{These authors have contributed equally to this work and share first authorship.}}\\Technion - IIT\\
    \And
    \href{https://orcid.org/0000-0002-5006-9300}
    {\includegraphics[scale=0.06]{orcid.pdf}\hspace{1mm}Ariel Larey$^{1,3,}$\footnotemark[1]}\\Technion - IIT\\
    \And
    \href{https://orcid.org/0000-0002-3855-299X}
    {\includegraphics[scale=0.06]{orcid.pdf}\hspace{1mm}Julie M. Caldwell$^{4,5}$}\\CCHMC\\
    \And
    \href{https://orcid.org/0000-0002-0756-5974}
    {\includegraphics[scale=0.06]{orcid.pdf}\hspace{1mm}Margaret H. Collins$^{5,6}$}\\CCHMC\\
    \And
    CEGIR Investigators Group$^{\#}$\\The Consortium of Eosinophilic\\Gastrointestinal Disease Researchers\\
    \And
    \href{https://orcid.org/0000-0001-9790-6332}
    {\includegraphics[scale=0.06]{orcid.pdf}\hspace{1mm}Marc E. Rothenberg$^{4,5}$}\\CCHMC\\
    \And
    \href{https://orcid.org/0000-0002-5345-8491}
    {\includegraphics[scale=0.06]{orcid.pdf}\hspace{1mm}Yonatan Savir$^{1,}$
    \thanks{Corresponding author, e-mail: yoni.savir@technion.ac.il. \newline
        $^{\#}$CEGIR members: Juan P. Abonia$^{4,5}$, Seema S. Aceves$^{7}$, Nicoleta C. Arva$^{8}$, Mirna Chehade$^{9}$, Evan S. Dellon$^{10}$, Nirmala Gonsalves$^{11}$, Sandeep K. Gupta$^{12}$, John Leung$^{13}$, Kathryn A. Peterson$^{14}$, Tetsuo Shoda$^{4,5}$, Jonathan M. Spergel$^{15}$.\newline
    $^{1}$Department of Physiology, Biophysics and System Biology, Faculty of Medicine, Technion Israel Institute of Technology, Haifa, Israel.
    $^{2}$Faculty of Industrial Engineering, Technion Israel Institute of Technology, Haifa, Israel.
    $^{3}$Faculty of Computer Science, Technion Israel Institute of Technology, Haifa, Israel. 
    $^{4}$Division of Allergy and Immunology, Cincinnati Children's Hospital Medical Center, Cincinnati, OH, USA.
    $^{5}$Dept. of Pediatrics, University of Cincinnati College of Medicine, Cincinnati, OH, USA.
    $^{6}$Division of Pathology and Laboratory Medicine, Cincinnati Children's Hospital Medical Center, Cincinnati, OH, USA.
    $^{7}$Division of Allergy and Immunology, University of California at San Diego, Rady Children's Hospital, San Diego, CA, USA.
    $^{8}$Dept. of Pathology and Laboratory, Ann and Robert H Lurie Children's Hospital, Northwestern University Feinberg School of Medicine, Chicago, IL, USA.
    $^{9}$Mount Sinai Center for Eosinophilic Disorders, Icahn School of Medicine at Mount Sinai, New York, NY, USA.
    $^{10}$Division of Gastroenterology and Hepatology, University of North Carolina School of Medicine, Chapel Hill, NC, USA.
    $^{11}$Division of Gastroenterology and Hepatology, Northwestern University Feinberg School of Medicine, Chicago, IL, USA.
    $^{12}$Division of Pediatric Gastroenterology, Hepatology and Nutrition, Children's of Alabama, University of Alabama at Birmingham, Birmingham, AL, USA.
    $^{13}$Boston Specialists, Boston, MA, USA.
    $^{14}$Division of Gastroenterology, University of Utah, Salt Lake City, UT, USA.
    $^{15}$Division of Allergy and Immunology, Children's Hospital of Philadelphia, Dept. of Pediatrics, Perelman School of Medicine at the University of Pennsylvania, Philadelphia, PA, USA.
}}\\Technion - IIT}

\renewcommand{\headeright}{Aknin et al.}
\renewcommand{\shorttitle}{EoE Dietary Treatment Assignment using Causal Learning}

\begin{document}

\maketitle

\begin{abstract}
Eosinophilic esophagitis (EoE) is a chronic, food antigen-driven, allergic inflammatory condition of the esophagus associated with elevated esophageal eosinophils. EoE is a top cause of chronic dysphagia after GERD. Diagnosis of EoE relies on counting eosinophils in histological slides, a manual and time-consuming task that limits the ability to extract complex patient-dependent features. The treatment of EoE includes medication and food elimination. A personalized food elimination plan is crucial for engagement and efficiency, but previous attempts failed to produce significant results. In this work, on the one hand, we utilize AI for inferring histological features from the entire biopsy slide, features that cannot be extracted manually. On the other hand, we develop causal learning models that can process this wealth of data. We applied our approach to the 'Six-Food vs. One-Food Eosinophilic Esophagitis Diet Study', where 112 symptomatic adults aged 18-60 years with active EoE were assigned to either a six-food elimination diet (6FED) or a one-food elimination diet (1FED) for six weeks. Our results show that the average treatment effect (ATE) of the 6FED treatment compared with the 1FED treatment is not significant, that is, neither diet was superior to the other. We examined several causal models and show that the best treatment strategy was obtained using T-learner with two XGBoost modules. While 1FED only and 6FED only provide improvement for 35\%-38\% of the patients, which is not significantly different from a random treatment assignment, our causal model yields a significantly better improvement rate of 58.4\%. This study illustrates the significance of AI in enhancing treatment planning by analyzing molecular features' distribution in histological slides through causal learning. Our approach can be harnessed for other conditions that rely on histology for diagnosis and treatment.
\end{abstract}


\section{Introduction}
Eosinophilic esophagitis (EoE) is characterized by eosinophil accumulation in the esophagus and is essentially caused as a result of food sensitivity. EoE is a clinicopathologic disorder, and its diagnosis relies on the assessment of patient symptoms by a physician which includes chest pain and dysphagia in adults, vomiting, failure to thrive, and abdominal pain in children \citep{Miehlke2015a}. In addition, patients undergo an esophagogastroduodenoscopy (EGD) during which several esophageal biopsies are procured. The tissue is fixed in formalin, processed and embedded in paraffin, sectioned onto slides, and subjected to hematoxylin and eosin (H\&E) staining \citep{Bancroft2008b}. A pathologist analyzes the biopsies to assess eosinophil infiltration and other histopathologic features such as abnormalities in the esophageal epithelium, lamina propria, and if present, muscularis mucosae. The gold standard for active EoE diagnosis requires that the individual’s esophageal biopsy exhibit a peak eosinophil count (PEC) of greater than or equal to 15 eosinophils per 400X high-power field (HFP) in the esophageal epithelium \citep{Dellon2010}. The PEC is identified by searching for the HPF in the whole slide image (WSI) that has the greatest number of eosinophils. Other microscopic features of EoE are assessed and quantified during pathology diagnosis by the EoE histology scoring system (EoEHSS) \citep{collins2017newly}, which assesses the severity and extent of eight features within the tissue. In recent works, we introduced an artificial intelligence (AI) system that predicts EoE activity by analyzing the eosinophil distribution within the entire WSI \citep{daniel2022deep, czyzewski2021machine}. An improved AI system predicts other EoE features and extracts more detailed information from the entire WSI \citep{Larey_2022}.

Currently, the main treatments are used to reduce the many discomforts that patients experience daily. They include diet, drugs, and esophageal dilation \citep{d2015eosinophilic}. Dietary therapy consists of at least three different possibilities: elemental diet, allergy test-based diet, and empiric elimination diet \citep{spergel2005treatment}. The elemental diet involves the use of amino-acid-based liquid formulas. Although this diet is highly effective (90\% response in children and 70\% in adults), it is infrequently accepted by patients because of its difficulty to implement. The allergy test-based diet uses a combination of skin prick and patch tests to identify trigger foods, followed by elimination of the identified foods from the diet; this diet shows a 70\% response rate for children but a low response rate for adults. Finally, empiric elimination diets may exclude up to six of the most allergenic foods, followed by gradual reintroduction of each food. Such diets show a reasonable response rate in adults \citep{lucendo2013empiric}. Having an improved method to identify the causative food(s) for an individual’s EoE would be greatly helpful because the success of a particular diet in inducing disease remission can only be assessed by endoscopy and biopsy, which has risks due to general anesthesia.

Drug treatment for EoE patients includes corticosteroids such as prednisolone or methylprednisolone. Studies demonstrated a high response rate for this treatment, but the recurrence of symptoms and eosinophilic infiltration is usually observed. Almost 40\% of patients with clinical EoE features respond to proton pump inhibitor (PPI) therapy, yet upon cessation of treatment, the disease recurs \citep{molina2015proton}. EoE patients usually develop esophageal strictures; therefore, endoscopic dilation can reduce pain and thus improve patients’ daily life, although the treatment does not reduce inflammation \citep{dellon2013acg}. 

Causal learning has become increasingly important in personalized treatment planning due to its ability to identify the causal relationships between different factors that contribute to a person's health outcomes. By understanding the causal mechanisms underlying a patient's condition, clinicians can develop personalized treatment plans that are tailored to the specific needs and circumstances of that individual \citep{pearl2016designing, van2021artificial, shalit2017estimating}. For example, comprehensive geriatric assessment (CGA) has been shown to improve the quality of life for older people with cancer who are starting systemic anti-cancer treatment \citep{soo2022integrated}. Another example is a method that aims to estimate the conditional average treatment effect (CATE) on disability progression in individuals with multiple sclerosis using a deep learning model \citep{falet2022estimating}. The model accurately predicts responders and non-responders to anti-CD20 antibodies, making it possible to adapt the treatment personally. In another paper, personalized treatment was used to reduce the incidence of obesity and obesity-related diseases \citep{chen2022personalized}. The researchers used meta-algorithms to estimate the personalized optimal decision on alcohol, vegetable, high-caloric food, and daily water intake respectively for each individual. The results showed that personalized treatment based on the meta-algorithms has better effectiveness to reduce obesity levels.

\begin{figure}[htbp]
\centerline{\includegraphics[width=\textwidth]{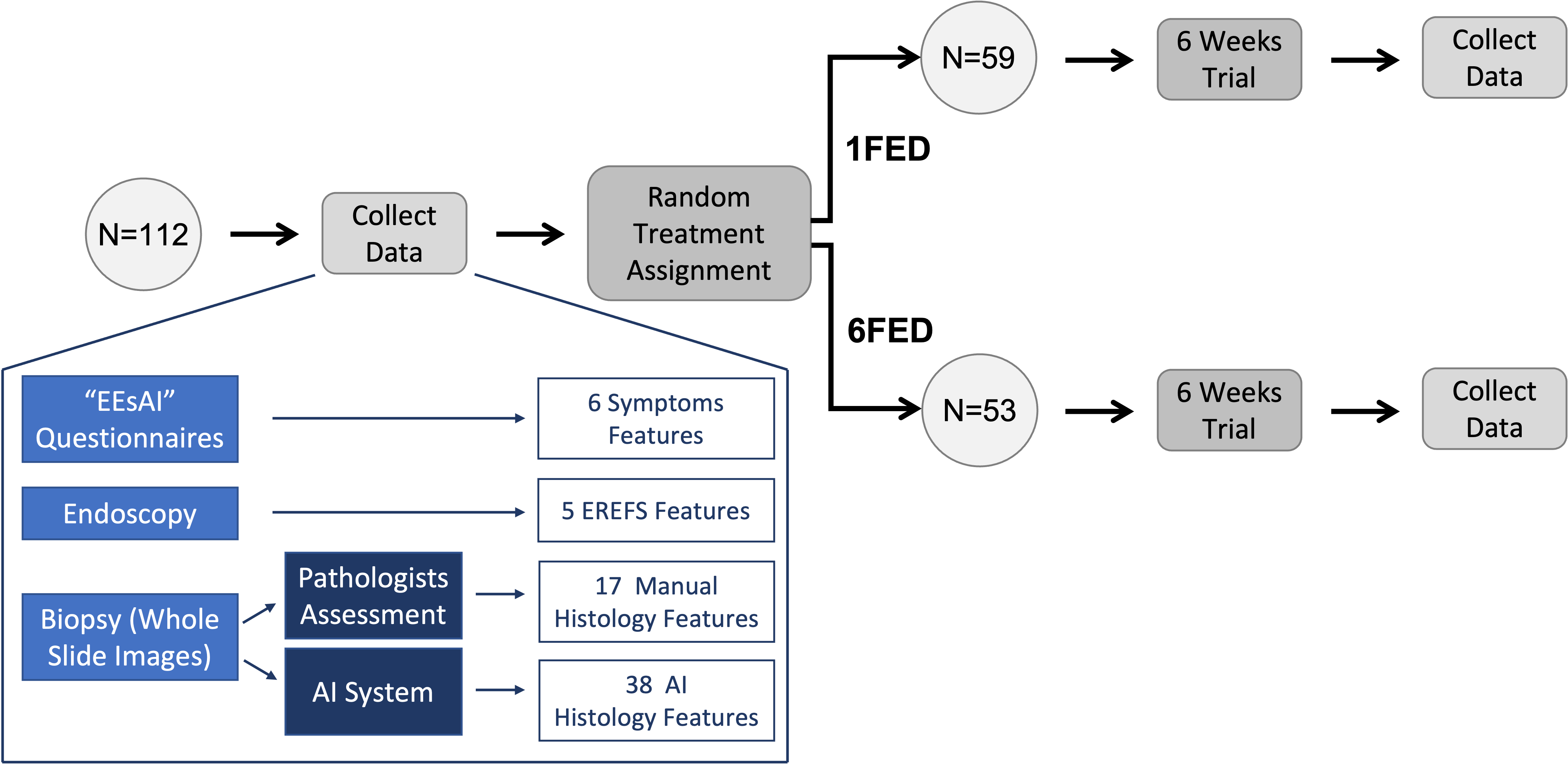}}
\caption{High-level trial profile. In this work, we focus on the first phase of the Six-Food vs. One-Food Eosinophilic Esophagitis Diet Study (SOFEED) trial. This study explored the effectiveness of two types of food elimination diets in treating eosinophilic esophagitis (EoE): a one-food elimination diet (1FED) and a six-food elimination diet (6FED). Only patients that completed the trial are included in our study. 112 patients were assigned randomly to one of the two treatments, where 59 patients were assigned to the 1FED, and the remaining 53 were assigned to 6FED. We utilize the information that was gathered from all patients at the beginning of the trial to predict treatment outcomes. This information included 66 features that were extracted from a patient-reported outcomes instrument (Eosinophilic Esophagitis Activity Index, EEsAI), endoscopic observations (EoE endoscopic reference score, EREFS), and histologic features of esophageal biopsy samples.
}
\label{Fig1}
\end{figure}

In this paper, our aim is to determine the diet treatment assignment most likely to induce disease remission for each individual EoE patient using information derived from their condition prior to the implementation of any diet. We use information from the ”Six-Food vs. One-Food Eosinophilic Esophagitis Diet Study” (SOFEED) randomized, open-label trial (Fig. \ref{Fig1}) to identify which diet is most suitable for each patient in order to increase the chance of inducing disease remission. Specifically, we examine the causal effect of these treatments on the clinical outcomes of PEC levels and EoE disease activity, each as determined by either pathologists or an AI system. 
Additionally, we utilize other EoE attributes, such as EoEHSS features, AI features, EEsAI symptoms questionnaire, and endoscopic reference score, as prior (pre-diet) information about the patients. We calculate the treatment effect of each diet when applied to all patients equally using the standard causal method of average treatment effect (ATE). Furthermore, we show how personalizing the treatments achieves an improved effect. The study was done using different methods and models and indicates that using machine learning models provides better performance than random assignment. 

\section{Methods}
\subsection{Six-Food vs. One-Food Eosinophilic Esophagitis Diet Study (SOFEED) Trial}
The ”Six-Food vs. One-Food Eosinophilic Esophagitis Diet Study” (SOFEED) was a multicenter, randomized, open-label trial that consisted of two phases \citep{gupta2019consortium, KLIEWER2023}. The first phase included the randomization of patients with active EoE to one of two diets (one-food elimination diet [1FED] or six-food elimination diet [6FED]). Patients who continued to the second phase were assigned to 6FED or topical swallowed steroid treatment. In this work, we focus only on the first phase of the trial. Patients with active EoE (PEC $\geq$ 15) were randomly assigned to one of the two food elimination diets: 1FED or 6FED. The 1FED excludes only animal milk, and the 6FED excludes animal milk, wheat, egg, soy, fish and shellfish, and peanut and tree nuts. This phase of the study lasted approximately six weeks. All participants who completed the first phase underwent an endoscopy to determine their disease status. Patients with PEC less than 15 were considered to be in remission, whereas patients having a PEC that is greater than or equal to 15 were considered to have active EoE and were assigned to a stricter treatment in the second phase (which is excluded from this study). Information from each patient used in this study was collected at randomization and six weeks. In the SOFEED study, 129 patients started the trial; 67 patients were assigned to the 1FED, and 62 patients were assigned to the 6FED. Due to missing information about patients because of withdrawal or missing biopsies, we excluded 17 patients from the dataset (8 that were assigned to 1FED and 9 that were assigned to 6FED). An illustration of the first phase of the trial process is described in Fig. \ref{Fig1}.

\subsection{Data Preparation}
The dataset for this study has four sources: endoscopic observations score (EREFS), manual assessment of histology (PEC, EoEHSS), AI prediction of histology (38 histology parameters), and Eosinophilic Esophagitis Activity Index (EEsAI) patient-reported outcome (PRO) scores (symptoms questionnaire).

\subsubsection{Endoscopy}
Each subject underwent endoscopy at the beginning of the trial and at the end of the first phase of the trial (six weeks). The presence and severity of the endoscopic findings of esophageal edema, rings, exudates, furrows, and stricture were assessed and reported as EoE endoscopic reference scores (EREFS)\citep{hirano2014role}.

\subsubsection{Histology – Manual Assessment}
Esophageal biopsies were procured from up to three locations (distal, proximal, middle) in the esophagus during the endoscopy; the biopsies were processed, embedded, sectioned, and H\&E-stained. Pathologists evaluated the slides, quantified the peak eosinophil count (PEC), and performed the EoEHSS to score the severity (grade) and extent (stage) of each of seven features \citep{collins2017newly}. In each endoscopy, more than one biopsy could be sampled (between one to three) from different locations in the esophagus. In this study, we utilize the information from the location that exhibited the most severe features to represent the patient’s histological features. 

\subsubsection{Histology – AI Prediction}
In recent work, we implemented an AI model that segments and evaluates different EoE features from the esophageal biopsy whole slide image \citep{daniel2022deep, Larey_2022}. We applied this system to extract additional 38 histological parameters. Several features aim to encapsulate the pathologists’ clinical methodologies (e.g., PEC), and several are novel metrics that we developed such as Spatial Eosinophil Count (SEC), Peak Basal Zone (PBZ), and Spatial Basal Zone (SBZ). For each endoscopy, the biopsy with the most severe quantification of a given feature was chosen to represent each patient’s histological features for downstream analyses. 

\subsubsection{Symptoms}
The Eosinophilic Esophagitis Activity Index (EEsAI) patient-reported outcome (PRO) instrument, which is a validated symptom diagnosis questionnaire for EoE patients, was performed by the patients at randomization and at the six-week timepoint, and the results were added to our dataset \citep{schoepfer2014international}. The features contained the response to questions regarding difficulties during eating or while swallowing food. 

\subsubsection{Data Features Properties}
In total, 59 patients were assigned to 1FED and 53 patients to 6FED. Each patient’s data sample contains 17 pathology features from the manual assessment, 38 AI-based features, 5 endoscopy features, and 6 symptom features (66 features in total). All features are severity features, which means that when their value is higher, the patient is experiencing a more severe condition or more extensive disease.

\subsubsection{Data Pre-Process}
The clinical outcomes are defined as the difference between the features at the end of the trial ($X_{end}$) and the initial features ($X_{start}$) before receiving the treatment. Formally:
\begin{equation}\label{eq1}
    Y=X_{end} - X_{start}
\end{equation}
To estimate the causal effects of the different outcomes, we use a common standardization (Z-score) approach:
\begin{equation}\label{eq2}
    z_{i,f}=\frac{y_{i,f}-mean(y_{f})}{std(y_{f})}
\end{equation}
Where $z_{i,f}$ is the Z score outcome of patient $i$ taken from feature $f$. $y_{i,f}$ is the original outcome of patient $i$ taken from feature $f$. $mean(y_{f})$ and $std(y_{f})$ are the average and standard-deviation of feature's $f$ outcomes calculated over all patients’ samples respectively.

 \begin{figure}[htbp]
\centerline{\includegraphics[width=\textwidth]{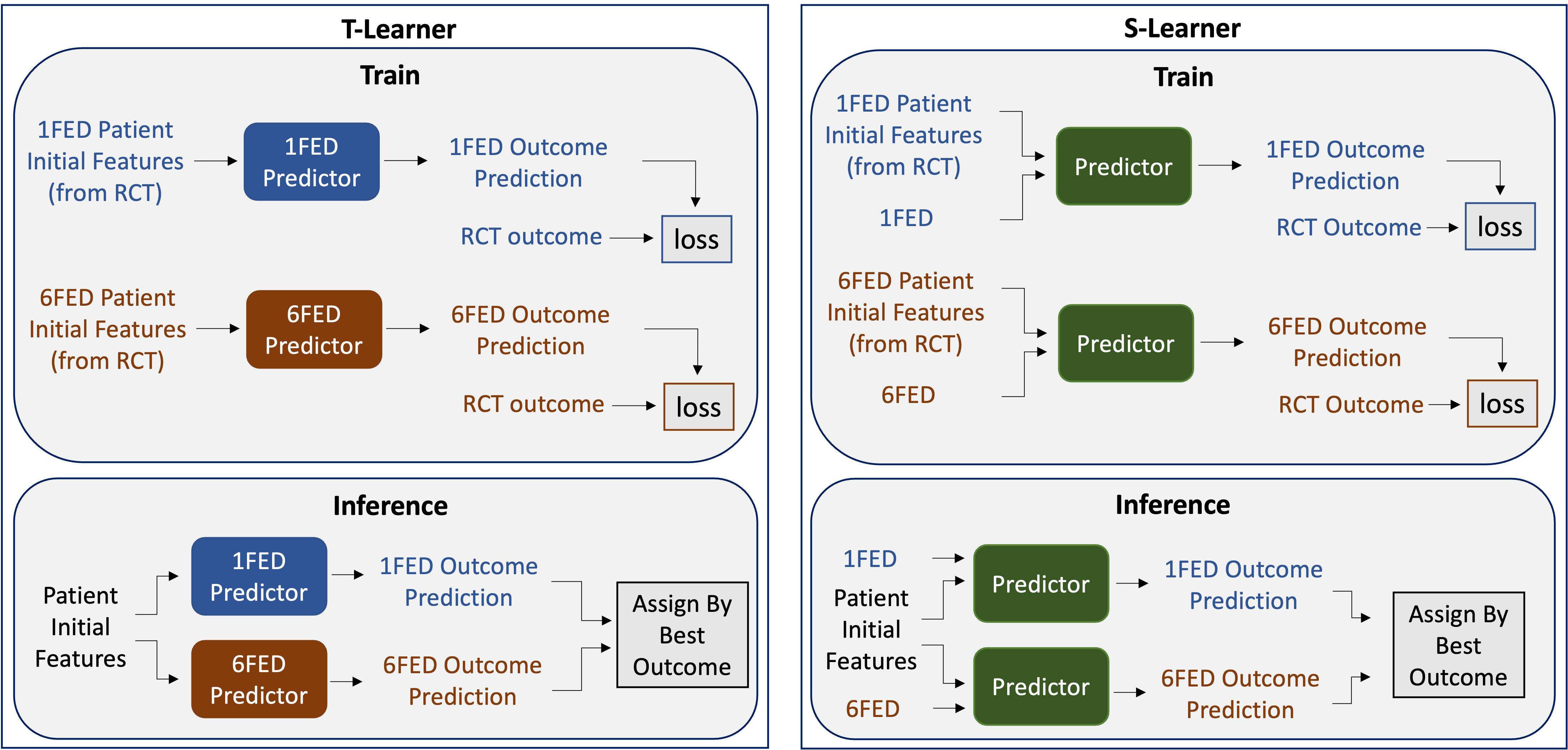}}
\caption{
Illustration of the two types of learners examined. T-learner is based on two different models where each model is dedicated to a different type of treatment and is trained separately, with its corresponding data, to predict the outcome. During inference, the patient's features are plugged into both models, and the treatment assignment is determined based on the model with the superior outcome. In the case of an S-learner, there is only one model that gets, in addition to the patients' data, the type of treatment that was given. During inference, the same model gets the treatment type as input. The treatment that results in a better outcome is the preferred one.
}
\label{Fig2}
\end{figure}

\subsection{Average Treatment Effects (ATE)}
In our case, as the treatment assignments are random, one can assume “Strong Ignorability”. That is, there are no confounders that influence treatment assignment, thus, the potential outcomes are independent of treatment assignment. Furthermore, “Stable Unit Treatment Value Assumption” (SUTVA) holds because an individual’s treatment does not influence another patient’s outcome. Moreover, because every patient has a probability to be assigned to every type of treatment (due to the random assignment), “Common Support” holds too.

The average treatment effect (ATE) \citep{twisk2018different} definition is:
\begin{equation}\label{eq3}
    ATE_{f} = E[z_{i,f,T=1} - z_{i,f,T=0}]
\end{equation}
Where the operator $E$ denotes expectation over all patients. $z_{i,f,T=j}$ is the outcome of the feature $f$ for the patient $i$ given the treatment $j$. In our case, the treatment is 1FED or 6FED. As a patient does not receive both treatments at the same time and because the causal identification assumptions hold we separate expectations and calculate ATE over the different outcomes via the next equation:
\begin{equation}\label{eq4}
    ATE_{f} = E[z_{i,f}|T=1] - E[z_{i,f}|T=0]
\end{equation}
For convenience, we use the following notation, T=1, refers to 6FED treatment and T=0 refers to 1FED treatment.

\subsection{Treatment Assignment Policy}
ATE represents the overall causal effects of the two treatments in different groups. This approach assumed the naïve policy for treatment assignment, where all patients were assigned the same treatment. To determine the best treatment strategy, we implemented a policy where individuals may be assigned to different treatments based on machine-learning model predictions. Particularly, for a given set of features $X_{start}$ (before trial), a model is trained to predict the outcomes for the two types of treatment assignments, where the real outcomes $z_{i,f}$ serves as the Ground Truth (GT). During inference, for a given patient’s $X_{start}$ features, the treatment is determined by the outcome with the superior effect. We examined two techniques \citep{kunzel2019metalearners} for this policy prediction (Fig. \ref{Fig2}): 
\begin{itemize}
    \item S-learner: Train a single model using the real treatment assignments to predict the outcomes, where the treatment $T$ is also an input to the model in addition to the features $X_{start}$. To predict treatment assignment, infer the trained model twice, once with treatment $T_{1}$ as input and once with $T_{6}$, and assign treatment based on the prediction with the better effect.
    \item T-learner: train two models; one is dedicated to $T_{1}$ and is trained only with the data of patients who were assigned to 1FED during the randomized trial, and the second model is dedicated to $T_{6}$ and was trained only with the data of patients who were assigned to 6FED. Each model is trained individually to predict the outcome given the features $X_{start}$. To predict treatment assignment, we infer $T_{1}$ model and $T_{6}$ model with the same patient’s $X_{start}$ features as an input and assign treatment based on the prediction with the superior effect.
\end{itemize}

\begin{figure}[htbp]
\centerline{\includegraphics[width=\textwidth]{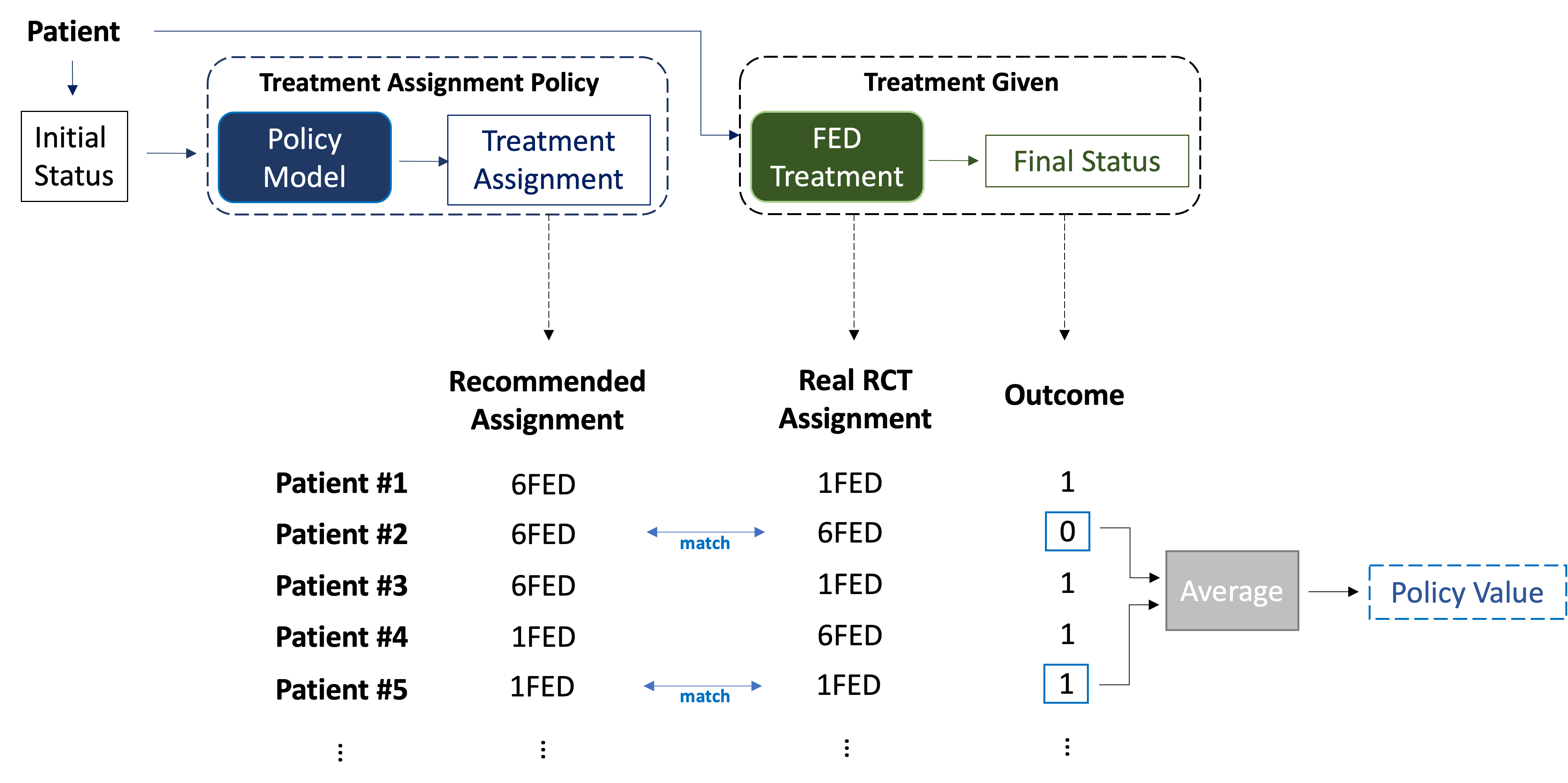}}
\caption{
Policy Value Calculation. Each patient is assigned to a treatment based on a given policy. The policy could be defined by a causal learner using the patient’s initial features, or it could be a naive policy that does not take the initial status of the patient before the trial into account. To evaluate the policy, we calculate the policy value by averaging the outcomes based on the samples that contain this information. More specifically, the outcomes that stem from the actual treatment assignments are equivalent to the given policy’s treatment assignments.
}
\label{Fig3}
\end{figure}

 For both learners, we trained various machine learning models. The first two are Decision Tree (DT) \citep{myles2004introduction} and XGBoost \citep{ramraj2016experimenting}. Both tree-based models can be trained even when the data have missing samples (i.e., some feature information is none), as occurs in our case. In addition, we trained Multi-Layer-Perceptron (MLP) \citep{gardner1998artificial} and Support vector machine (SVM) \citep{hearst1998support} as well. In these cases, the missing samples were filled with the average value of the corresponding feature in the training set. We trained both learners with different architectures and various hyperparameters and reported the best results. Each training process was performed with a 6-fold cross-validation (CV) \citep{refaeilzadeh2009cross}, wherein each CV iteration we train the model for policy prediction using a training set and assign treatment on the validation set based on the trained model. By this technique, we eventually yield a treatment assignment prediction to the model over the entire dataset, while each prediction was achieved by a model that was not trained on the observed sample.

\begin{table}[ht]
\footnotesize
\caption{Average treatment effects (ATE) results for peak eosinophil count (PEC) and EoE-Active outcomes.}
\label{tab:table1} 
\centering
\begin{tabular}{|c||c|c|c|c|}
  \hline
    {} & {PEC-AI} & {PEC-Manual} & {EoE Active-AI} &  {EoE Active-Manual} \\
  \hline
  \hline
    ATE & -13.55 & -9.03 & 0.05 & 0.02 \\
  \hline
    CI - 95\% & [-46.45, 15.11] & [-27.09, 9.92] & [-0.16, 0.25]  & [-0.16, 0.19] \\
  \hline
    P-value & 0.202 & 0.168 & 0.318 & 0.405 \\
  \hline
\end{tabular}
\end{table}

\subsection{Policy Value}
The policy value is the average outcome calculated over the patients that were assigned to treatment based on a given policy. Since the subject received only one of two treatments, we calculate the policy value by referring only to the subjects whose treatment assignment was identical to the actual randomized trial assignment, enabling access to the corresponding outcome values (Fig. \ref{Fig3}). We calculated the policy value for different treatment assignment policies and compared them:
\begin{itemize}
    \item ML Policy: A policy that is predicted by a learner and is based on machine learning models. We trained S-learner and T-learner over different models and evaluated them based on their policy values.
    \item PEC Policy: A naive policy that is based on PEC only, which is the gold-standard histologic metric for diagnosing EoE. We examined all possible different thresholds of PEC values at the randomization timepoint of the trial ($X_{start}$) and determined the treatment assignment for each side of the threshold.
    \item 6FED Policy: A baseline policy where all patients receive the 6FED treatment regardless of their initial condition.
    \item 1FED Policy: A baseline policy where all patients receive the 1FED treatment regardless of their initial condition.
    \item Random Policy: A baseline policy where each patient treatment is assigned randomly. We performed this policy assignment 1,000 times with different random seeds and reported the policy value statistics.
\end{itemize}

\begin{table*}[ht]
\footnotesize
\caption{Diverse Policy Values for PEC Reduction and Effectiveness.}
\centering
\begin{tabular}{|c||c|c||c|c|}
  \hline
    Method & PEC-AI Decrease & PEC-Manual Decrease & Effectiveness-AI & Effectiveness-Manual \\
  \hline
  \hline
  1FED Policy & 27.2 & 25.1 & 27.1\% & 35.6\% \\
  \hline
  6FED Policy & 14.8 & 15.9 & 32.1\% & 37.7\% \\
  \hline
  Random Policy & 21.4 & 20.4 & 29.4\% & 36.7\% \\
  CI 95\% & [9.1, 34.7] & [12.5, 28.6] & [20.7\%, 38.2\%]  & [28.5\%, 44.2\%]\\
    \hline
  PEC-AI Baseline & 22.3 & 24.7 & 30.4\% & 36.8\% \\
    \hline
  PEC-Manual Baseline & 24.5 & 18.2 & 30.9\% & 38.3\% \\
    \hline
  S-Learner Policy: & {} & {} & {} & {} \\
  MLP & 38.3 & 32.5 & 41.4\% & 49\% \\
  XGBoost & 35.5 & 29.7 & 32.1\% & 41.3\% \\
      \hline
  T-Learner Policy: & {} & {} & {} & {} \\
  MLP & 45.2 & 36.5 & 46.4\% & 50.1\% \\
  XGBoost & 45.9 & 39.9 & 52\% & 58.4\% \\
  \hline
\end{tabular}
\label{tab:table2} 
\end{table*}

\section{Results}
\subsection{Overall Treatment Assignment}
This study considers two types of outcomes. The first outcome metric is the decrease in the PEC score due to the treatment, which is a continuous metric. The second outcome metric is treatment effectiveness - the change in the patient activity (that is whether PEC is lower or above 15), which is a binary metric. The average treatment effectiveness over all the patients is the percentage of patients that responded to the treatment.

For example, a treatment that reduces the PEC score from 80 eosinophils to 60 eosinophils would have a PEC decrease of 20 but without a change in activity. In comparison, a treatment that reduces the PEC score from 20 eosinophils to 10 eosinophils would have a PEC decrease of 10 eosinophils and a change in activity.
We use two methods to assess the PEC score of the patient (which determines both outcome metrics). The first is using the pathologist's manual assessment and the second is using the AI-based automated assessment \citep{daniel2022deep, Larey_2022}. Overall, we consider four metrics for the treatment outcome: PEC-AI decrease, PEC-manual decrease, Effectiveness-AI, and Effectiveness-manual

We calculated the ATE for these four outcomes (Table \ref{tab:table1}). The results show that the treatments reduce the EoE activity but in all cases, there is no significant advantage in assigning one treatment over the other. That is, assigning 6FED all the time has no advantage over assigning 1FED all the time (and vice versa), in terms of ATE.

\begin{figure}[htbp]
\centerline{\includegraphics[width=\textwidth, trim=3 2 2 1, clip]{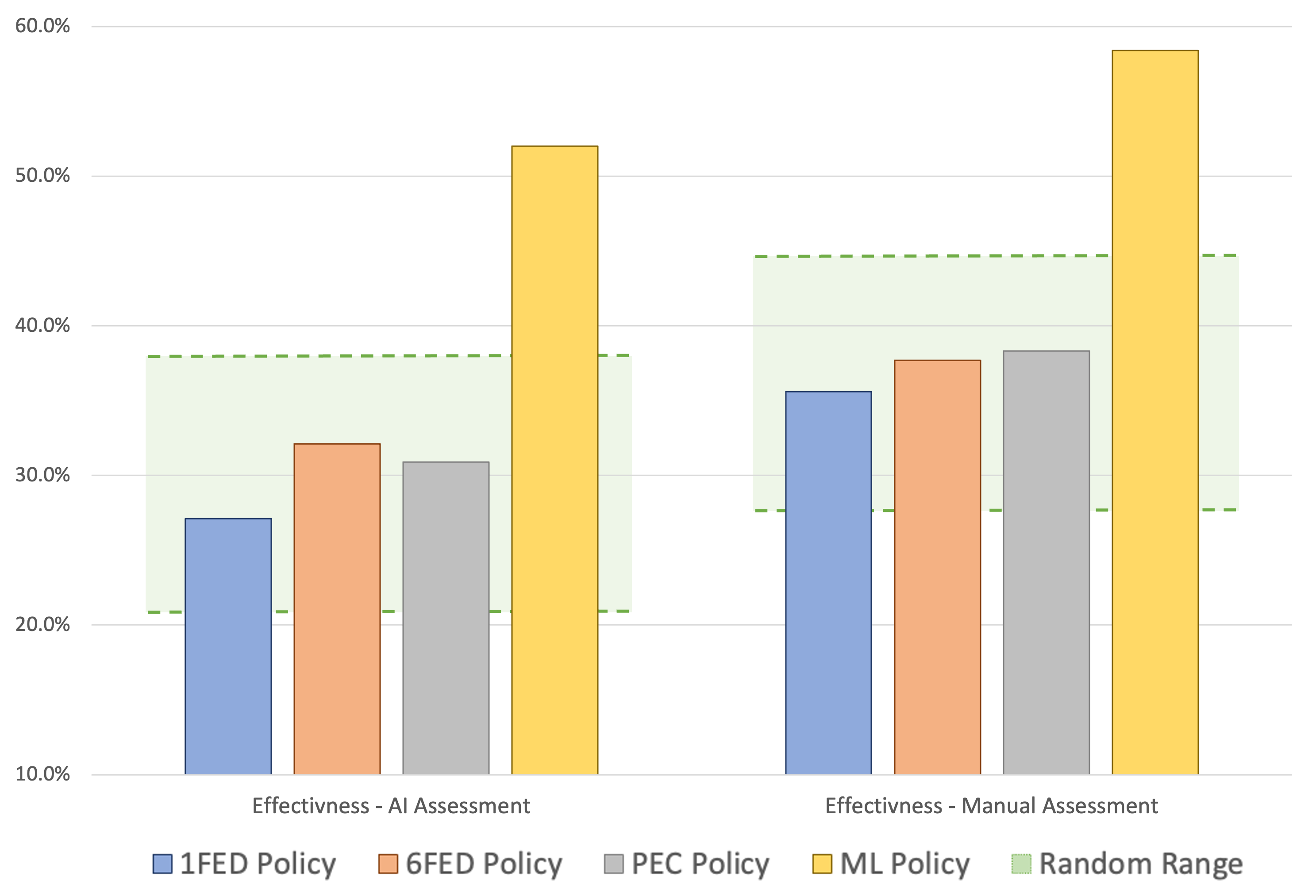}}
\caption{The treatment effectiveness, the percentage of patients that respond to the treatment for different policies. The graph compares four different policies. 1FED policy (blue) and 6FED policy (orange) are the naive approaches where all patients are assigned to the same treatment. The PEC policy (grey) is based only on the initial PEC (at the beginning of the trial), where patients with high PEC were assigned 6FED and the rest by 1FED. The causal (ML) policy (yellow) is the trained T-learner model. The PEC that characterizes the patients' outcome can be determined using manual assessment or using an automated AI platform (right and left, respectively). The green-shaded area is the distribution (95\% CI) of the effectiveness of a random policy where 1FED or 6FED are assigned randomly. All policies overlap with the random policy confidence interval, except for our T-learner which achieved significantly better (pval$<$0.001) performances.}
\label{Fig4}
\end{figure}

\subsection{Personalized Treatment Assignment}
Next, we trained the different ML models, performed the learners, and evaluated them by calculating their policy values. Our results show that the T-learner model (which consists of two XGBoost sub-modules) has the best performance in all four types of outcomes (Table \ref{tab:table2}). 
To evaluate the significance of any policy, it is crucial to estimate its performance with the effect of a random policy. Assigning treatments in a random fashion results in treatment effectiveness distribution with 95\% confidence intervals of [20.7\% 38.2\%] and [28.5\% 44.2\%], using AI-based PEC or manual PEC, respectively (Table \ref{tab:table2}, Fig. \ref{Fig4}). 1FED only, 6FED only, and PEC-based policy provide treatment effectiveness that is not significantly better that a random assignment (Fig. \ref{Fig4}). However, our ML policy provides treatment effectiveness of 52\%, in the case of AI-based PEC, and 58.4\% in the case of manual-based PEC. That is, our results provide a policy strategy that yields a significantly better (p-val$<$0.001) treatment effectiveness.

\subsection{Important Features}
\begin{figure}[htbp]
\centerline{\includegraphics[width=\textwidth]{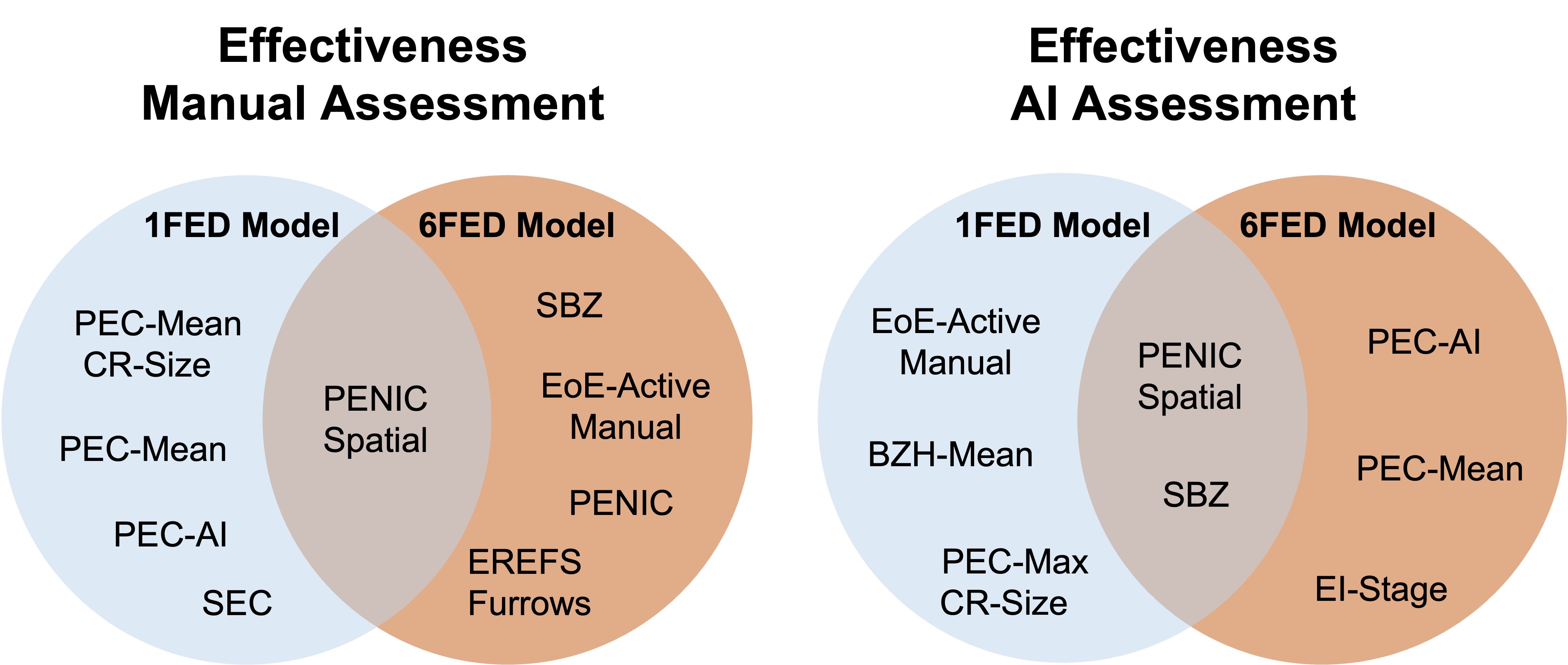}}
\caption{Top-5 significant features of the T-learner modules. The main features that are significant in all cases are the spatial distribution of regions with high eosinophil not-intact count ("PENIC spatial"), and the density of basal zone region in the whole slide image (SBZ). Abbreviation legend:
PEC - Peak Eosinophil Count,
SEC - Spatial Eosinophil Count,
PENIC - Peak Eosinophil Not-Intact Count,
SBZ - Spatial Bazal Zone,
CR - Connected Region,
EI - Esophageal Inflammation.
See results section (C) for a detailed description.}
\label{Fig5}
\end{figure}

The best casual model is the T-learner, which consists of two models; one is dedicated to the 1FED outcome, and the second to the 6FED. We found that the best models were based on XGBoost (Table \ref{tab:table2}), which consists of a features attention mechanism that predicts the input features gain and indicates their importance. To gain insight into the histological markers that are significant for the treatment assignment, we compared the significant features of the 1FED and 6FED models, whether the PEC is assessed manually or is AI-based (Fig. \ref{Fig5}). 

The only feature that is significant in all the cases is a histological feature that is a proxy to the spatial distribution of regions with high eosinophil not-intact count ("PENIC spatial"). Another common feature measures the density of the basal zone region in the whole slide image (SBZ). Interestingly, previous studies demonstrated a significant correlation between PEC and SBZ, and that SBZ is informative in predicting remission \citep{Larey_2022}.

Other histological features had high attention in the ML model prediction: Spatial Eosinophil Count (SEC) is a score that represents the distribution of eosinophils' appearance within the slide, whereas peak eosinophil count (PEC) is a local representation of the densest region. PEC-Mean and BZH-Mean are the average scores of the high-power fields (HPFs) examined over the biopsy slide regarding eosinophils and basal cells respectively. A connected region (CR) is a group of neighboring HPFs that each one of which contains eosinophils. Each CR size is measured as the number of HPFs within it. PEC-Mean-CR-size and PEC-Max-CR-Size represent the average size of the CRs in the slide and their maximal size respectively.
Not all important features were based on AI; a few metrics assessed manually were important as well, such as EI-Stage and EREFS-Furrows. The EoEHSS score of EI-Stage is assessed manually by the pathologist and represents the distribution of the eosinophils within the slide. EREFS-Furrows is a score that is based on the appearance of vertical lines in the esophagus and is measured during the endoscopy.

\section{Conclusions}
Personalized treatment planning requires comprehensive patient information and the tools to analyze and gain insights from the data. EoE is one of many conditions, such as many types of cancer, that its diagnosis and treatment planning relies heavily on histological examination. Traditional manual assessment of whole histological slides is a laborious, time-consuming, and somewhat subjective task and as a result, the number of features that determine the diagnosis is often limited. This is also the case for EoE where the gold standard for EoE diagnosis is the peak eosinophil count. One of the promises of AI in digital pathology, besides automating manual tasks, is the ability to process the entire WSI and infer a large number of histological markers that can provide the "histological fingerprint" that can facilitate personalized decision-making. While in previous studies we developed the AI platform that extracts histological biomarker that improves diagnosis \citep{daniel2022deep, Larey_2022}, here we show that using these features and causal learning improve treatment planning.

Diet is an important part of the treatment plan for EoE, as many patients have food triggers that can exacerbate symptoms. Choosing how strict the elimination should be for each patient is critical to treatment success. In this study, we examined the effect of two types of diet treatments on EoE patients. We used the data from the SOFEED trial that included endoscopy scores, manual histological scores, and symptom information. We further enlarged the dataset by applying the tissue WSIs to our AI system, achieving spatial and local information on different EoE features within the tissue.

The average treatment effect (ATE) analysis of the two different diets provided in the trial did not show any significant preference for one treatment over the other and is consistent with previous findings \citep{KLIEWER2023}. Assigning the same treatment to all patients yields effectiveness, that is the percentage of the patients that respond to the treatment of 35.6\% and 37.7\% for 1FED and 6FED treatments, respectively. The policy where the treatment assignment is based on PEC only gives an effectiveness of 38.3\%. Importantly, all these three approaches do not provide significantly better results compared to a random diet assessment. A random diet treatment results in an average effectiveness of 36.7\% with 95\% confidence intervals of [28.5\% 44.2\%]. Our T-learner provides a significant improvement in treatment with an effectiveness of 58.4\%

Our results show that the T-learner had better performance than S-learner in all outcomes. It emphasizes the importance of training two different models dedicated to each treatment individually, rather than training one model dedicated to both treatments and distinguished only by a negligible treatment feature as an input. Moreover, XGBoost showed better results than the other machine learning models regarding all four clinical outcomes as well.

When studying our best model we can identify the important features that were taken into account to train and learn the models. The most significant features for the treatment assignment are not manual histological features but rather features that were extracted from our digital pathology AI system. The most common one that was relevant in all models was based on the spatial distribution of not-intact eosinophils and was extracted from AI as well. An additional important feature extracted by the AI tool is the spatial distribution of basal cells, this feature's importance has been previously demonstrated in previous work \citep{Larey_2022}. Interestingly, the most important features involve measures of density over the entire WSI rather than local features which is consistent with the observation that their global, spatial, and histological features contribute to the EoE diagnosis \citep{czyzewski2021machine}.

This work highlights the importance of systematically analyzing the distribution of biopsy features over the entire slide image and integrating them together with casual learning to provide better treatment planning. Our approach can be used for other conditions that rely on histology for diagnosis and treatment.

\section*{Acknowledgment}
We thank Tanya Wasserman for fruitful discussions. This work is supported by the Cincinnati Children’s-Technion Bridge to Next Gen Medicine Grant.

\bibliographystyle{plainnat}

\begin{thebibliography}{}

\bibitem{Miehlke2015a}
S.~Miehlke, ``{Clinical features of Eosinophilic esophagitis in children and
  adults},'' \emph{Best Practice {\&} Research Clinical Gastroenterology},
  vol.~29, no.~5, pp. 739--748, 2015.

\bibitem{Bancroft2008b}
J.~D. Bancroft and M.~Gamble, \emph{{Theory and practice of histological
  techniques}}.\hskip 1em plus 0.5em minus 0.4em\relax Elsevier health
  sciences, 2008.

\bibitem{Dellon2010}
E.~S. Dellon, K.~J. Fritchie, T.~C. Rubinas, J.~T. Woosley, and N.~J. Shaheen,
  ``{Inter- and intraobserver reliability and validation of a new method for
  determination of eosinophil counts in patients with esophageal
  eosinophilia},'' \emph{Digestive Diseases and Sciences}, vol.~55, no.~7, pp.
  1940--1949, jul 2010.

\bibitem{collins2017newly}
M.~H. Collins, L.~J. Martin, E.~S. Alexander, J.~T. Boyd, R.~Sheridan, H.~He,
  S.~Pentiuk, P.~Putnam, J.~Abonia, V.~Mukkada \emph{et~al.}, ``Newly developed
  and validated eosinophilic esophagitis histology scoring system and evidence
  that it outperforms peak eosinophil count for disease diagnosis and
  monitoring,'' \emph{Diseases of the Esophagus}, vol.~30, no.~3, p.~1, 2017.

\bibitem{daniel2022deep}
N.~Daniel, A.~Larey, E.~Aknin, G.~A. Osswald, J.~M. Caldwell, M.~Rochman, M.~H.
  Collins, G.-Y. Yang, N.~C. Arva, K.~E. Capocelli \emph{et~al.}, ``A deep
  multi-label segmentation network for eosinophilic esophagitis whole slide
  biopsy diagnostics,'' in \emph{2022 44th Annual International Conference of
  the IEEE Engineering in Medicine \& Biology Society (EMBC)}.\hskip 1em plus
  0.5em minus 0.4em\relax IEEE, 2022, pp. 3211--3217.

\bibitem{czyzewski2021machine}
T.~Czyzewski, N.~Daniel, M.~Rochman, J.~M. Caldwell, G.~A. Osswald, M.~H.
  Collins, M.~E. Rothenberg, and Y.~Savir, ``Machine learning approach for
  biopsy-based identification of eosinophilic esophagitis reveals importance of
  global features,'' \emph{IEEE open journal of engineering in medicine and
  biology}, vol.~2, pp. 218--223, 2021.

\bibitem{Larey_2022}
A.~Larey, E.~Aknin, N.~Daniel, G.~A. Osswald, J.~M. Caldwell, M.~Rochman,
  T.~Wasserman, M.~H. Collins, N.~C. Arva, G.-Y. Yang, M.~E. Rothenberg, and
  Y.~Savir, ``Harnessing artificial intelligence to infer novel spatial
  biomarkers for the diagnosis of eosinophilic esophagitis,'' \emph{Frontiers
  in Medicine}, vol.~9, oct 2022.

\bibitem{d2015eosinophilic}
A.~D’Alessandro, D.~Esposito, M.~Pesce, R.~Cuomo, G.~D. De~Palma, and
  G.~Sarnelli, ``Eosinophilic esophagitis: from pathophysiology to treatment,''
  \emph{World journal of gastrointestinal pathophysiology}, vol.~6, no.~4, p.
  150, 2015.

\bibitem{spergel2005treatment}
J.~M. Spergel, T.~Andrews, T.~F. Brown-Whitehorn, J.~L. Beausoleil, and C.~A.
  Liacouras, ``Treatment of eosinophilic esophagitis with specific food
  elimination diet directed by a combination of skin prick and patch tests,''
  \emph{Annals of Allergy, Asthma \& Immunology}, vol.~95, no.~4, pp. 336--343,
  2005.

\bibitem{lucendo2013empiric}
A.~J. Lucendo, {\'A}.~Arias, J.~Gonz{\'a}lez-Cervera, J.~L. Yag{\"u}e-Compadre,
  D.~Guagnozzi, T.~Angueira, S.~Jim{\'e}nez-Contreras,
  S.~Gonz{\'a}lez-Castillo, B.~Rodr{\'\i}guez-Dom{\'\i}ngez, L.~C. De~Rezende
  \emph{et~al.}, ``Empiric 6-food elimination diet induced and maintained
  prolonged remission in patients with adult eosinophilic esophagitis: a
  prospective study on the food cause of the disease,'' \emph{Journal of
  Allergy and Clinical Immunology}, vol. 131, no.~3, pp. 797--804, 2013.

\bibitem{molina2015proton}
J.~Molina-Infante, D.~A. Katzka, and E.~S. Dellon, ``Proton pump
  inhibitor-responsive esophageal eosinophilia: a historical perspective on a
  novel and evolving entity.'' \emph{Revista espanola de enfermedades
  digestivas: organo oficial de la Sociedad Espanola de Patologia Digestiva},
  vol. 107, no.~1, pp. 29--36, 2015.

\bibitem{dellon2013acg}
E.~S. Dellon, N.~Gonsalves, I.~Hirano, G.~T. Furuta, C.~A. Liacouras, and D.~A.
  Katzka, ``Acg clinical guideline: evidenced based approach to the diagnosis
  and management of esophageal eosinophilia and eosinophilic esophagitis
  (eoe),'' \emph{Official journal of the American College of Gastroenterology|
  ACG}, vol. 108, no.~5, pp. 679--692, 2013.

\bibitem{pearl2016designing}
C.~Pearl, \emph{Designing voice user interfaces: Principles of conversational
  experiences}.\hskip 1em plus 0.5em minus 0.4em\relax " O'Reilly Media, Inc.",
  2016.

\bibitem{van2021artificial}
M.~Van~der Schaar, A.~M. Alaa, A.~Floto, A.~Gimson, S.~Scholtes, A.~Wood,
  E.~McKinney, D.~Jarrett, P.~Lio, and A.~Ercole, ``How artificial intelligence
  and machine learning can help healthcare systems respond to covid-19,''
  \emph{Machine Learning}, vol. 110, pp. 1--14, 2021.

\bibitem{shalit2017estimating}
U.~Shalit, F.~D. Johansson, and D.~Sontag, ``Estimating individual treatment
  effect: generalization bounds and algorithms,'' in \emph{International
  Conference on Machine Learning}.\hskip 1em plus 0.5em minus 0.4em\relax PMLR,
  2017, pp. 3076--3085.

\bibitem{soo2022integrated}
W.~K. Soo, M.~T. King, A.~Pope, P.~Parente, P.~D{\=a}rzi{\c{n}}{\v{s}}, and
  I.~D. Davis, ``Integrated geriatric assessment and treatment effectiveness
  (integerate) in older people with cancer starting systemic anticancer
  treatment in australia: a multicentre, open-label, randomised controlled
  trial,'' \emph{The Lancet Healthy Longevity}, vol.~3, no.~9, pp. e617--e627,
  2022.

\bibitem{falet2022estimating}
J.-P.~R. Falet, J.~Durso-Finley, B.~Nichyporuk, J.~Schroeter, F.~Bovis, M.-P.
  Sormani, D.~Precup, T.~Arbel, and D.~L. Arnold, ``Estimating individual
  treatment effect on disability progression in multiple sclerosis using deep
  learning,'' \emph{Nature communications}, vol.~13, no.~1, pp. 1--12, 2022.

\bibitem{chen2022personalized}
S.~Chen, Y.~Dai, X.~Ma, H.~Peng, D.~Wang, and Y.~Wang, ``Personalized optimal
  nutrition lifestyle for self obesity management using metaalgorithms,''
  \emph{Scientific Reports}, vol.~12, no.~1, pp. 1--12, 2022.

\bibitem{gupta2019consortium}
S.~K. Gupta, G.~W. Falk, S.~S. Aceves, M.~Chehade, M.~H. Collins, E.~S. Dellon,
  N.~Gonsalves, I.~Hirano, V.~A. Mukkuda, K.~A. Peterson \emph{et~al.},
  ``Consortium of eosinophilic gastrointestinal disease researchers: advancing
  the field of eosinophilic gi disorders through collaboration,''
  \emph{Gastroenterology}, vol. 156, no.~4, pp. 838--842, 2019.

\bibitem{KLIEWER2023}
K.~L. Kliewer, N.~Gonsalves, E.~S. Dellon, D.~A. Katzka, J.~P. Abonia, S.~S.
  Aceves, N.~C. Arva, J.~A. Besse, P.~A. Bonis, J.~M. Caldwell \emph{et~al.},
  ``One-food versus six-food elimination diet therapy for the treatment of
  eosinophilic oesophagitis: a multicentre, randomised, open-label trial,''
  \emph{The Lancet Gastroenterology \& Hepatology}, 2023.

\bibitem{hirano2014role}
I.~Hirano, ``Role of advanced diagnostics for eosinophilic esophagitis,''
  \emph{Digestive Diseases}, vol.~32, no. 1-2, pp. 78--83, 2014.

\bibitem{schoepfer2014international}
A.~Schoepfer, A.~Straumann, R.~Panczak, M.~Coslovsky, C.~Kuehni, E.~Maurer,
  N.~Haas, Y.~Romero, I.~Hirano, J.~Alexander \emph{et~al.}, ``International
  eosinophilic esophagitis activity index study g. development and validation
  of a symptom-based activity index for adults with eosinophilic esophagitis,''
  \emph{Gastroenterology}, vol. 147, no.~6, pp. 1255--1266, 2014.

\bibitem{twisk2018different}
J.~Twisk, L.~Bosman, T.~Hoekstra, J.~Rijnhart, M.~Welten, and M.~Heymans,
  ``Different ways to estimate treatment effects in randomised controlled
  trials,'' \emph{Contemporary clinical trials communications}, vol.~10, pp.
  80--85, 2018.

\bibitem{kunzel2019metalearners}
S.~R. K{\"u}nzel, J.~S. Sekhon, P.~J. Bickel, and B.~Yu, ``Metalearners for
  estimating heterogeneous treatment effects using machine learning,''
  \emph{Proceedings of the national academy of sciences}, vol. 116, no.~10, pp.
  4156--4165, 2019.

\bibitem{myles2004introduction}
A.~J. Myles, R.~N. Feudale, Y.~Liu, N.~A. Woody, and S.~D. Brown, ``An
  introduction to decision tree modeling,'' \emph{Journal of Chemometrics: A
  Journal of the Chemometrics Society}, vol.~18, no.~6, pp. 275--285, 2004.

\bibitem{ramraj2016experimenting}
S.~Ramraj, N.~Uzir, R.~Sunil, and S.~Banerjee, ``Experimenting xgboost
  algorithm for prediction and classification of different datasets,''
  \emph{International Journal of Control Theory and Applications}, vol.~9,
  no.~40, 2016.

\bibitem{gardner1998artificial}
M.~W. Gardner and S.~Dorling, ``Artificial neural networks (the multilayer
  perceptron)—a review of applications in the atmospheric sciences,''
  \emph{Atmospheric environment}, vol.~32, no. 14-15, pp. 2627--2636, 1998.

\bibitem{hearst1998support}
M.~A. Hearst, S.~T. Dumais, E.~Osuna, J.~Platt, and B.~Scholkopf, ``Support
  vector machines,'' \emph{IEEE Intelligent Systems and their applications},
  vol.~13, no.~4, pp. 18--28, 1998.

\bibitem{refaeilzadeh2009cross}
P.~Refaeilzadeh, L.~Tang, and H.~Liu, ``Cross-validation.'' \emph{Encyclopedia
  of database systems}, vol.~5, pp. 532--538, 2009.

\end{thebibliography}

\end{document}